\documentclass[hidelinks]{svproc}
\usepackage{url}

\usepackage{hyperref}
\usepackage{breakurl}
\usepackage[backend=biber, style=numeric, sorting=none, hyperref, maxnames=70, minnames=4,
giveninits=true, maxcitenames=2, mincitenames=1, bibencoding=utf8, isbn=false, eprint=true]{biblatex}
\usepackage{mathptmx}
\usepackage{amsmath}
\usepackage{amstext}
\usepackage{amssymb}
\usepackage{mathtools}
\DeclarePairedDelimiter{\abs}{\lvert}{\rvert}
\usepackage{csquotes}

\usepackage[usenames,dvipsnames]{xcolor}
\usepackage{xspace}
\usepackage{todonotes}
\colorlet{fhcolor}{ProcessBlue}

\usepackage[capitalize]{cleveref}
\usepackage{scalefnt}

\usepackage{xpatch}
\xpatchbibdriver{article}
  {\usebibmacro{doi+url}}
  {\usebibmacro{doi+url}%
   \newunit\newblock
   \usebibmacro{eprint}}
  {}
  {}

\usepackage{algorithm}
\usepackage[]{algpseudocode}

\usepackage{listings}
\definecolor{dark_green}{rgb}{0,0.5,0}

\usepackage{subcaption}
\usepackage{tikz}
\usetikzlibrary{patterns}
\usetikzlibrary{decorations.pathreplacing}
\usetikzlibrary{decorations.pathmorphing}
\usetikzlibrary{positioning}
\usetikzlibrary{shapes.geometric}
\usetikzlibrary{arrows.meta}

\usepackage[bottom]{footmisc}
\usepackage{pifont}
\usepackage{makecell}

\begin{document}
\mainmatter              
\title{OpTC -- A Toolchain for Deployment of Neural Networks on AURIX TC3xx Microcontrollers}
\titlerunning{OpTC -- A Toolchain for Deployment of Neural Networks} 
\author{Christian Heidorn\inst{1} \and Frank Hannig\inst{1} \and 
Dominik Riedelbauch\inst{2} \and \\ Christoph Strohmeyer\inst{2} \and Jürgen Teich\inst{1}}
\authorrunning{Heidorn et al.} 
%
\tocauthor{Christian Heidorn, Frank Hannig, Dominik Riedelbauch, Christoph Strohmeyer, and Jürgen Teich}
\institute{Department of Computer Science,\\ Friedrich-Alexander-Universität Erlangen-Nürnberg (FAU), Germany\\ \email{christian.heidorn@fau.de}
\and Schaeffler Technologies AG \& Co. KG, Herzogenaurach, Germany}

\maketitle              

\begin{abstract}
The AURIX 2xx and 3xx families of TriCore microcontrollers are widely used in the automotive industry and, recently, also in applications that involve machine learning tasks.
Yet, these applications are mainly engineered manually, and only little tool support exists for bringing neural networks to TriCore microcontrollers.
Thus, we propose OpTC, an end-to-end toolchain for automatic compression, conversion, code generation, and deployment of neural networks on TC3xx microcontrollers.
OpTC supports various types of neural networks and provides compression using layer-wise pruning based on sensitivity analysis for a given neural network.
The flexibility in supporting different types of neural networks, such as multi-layer perceptrons (MLP), convolutional neural networks (CNN), and recurrent neural networks (RNN), is shown in case studies for a TC387 microcontroller. Automotive applications for predicting the temperature in electric motors and detecting anomalies are thereby used to demonstrate the effectiveness and the wide range of applications supported by OpTC.
\end{abstract}

\section{Introduction}
\label{sec:introduction}

With the \textit{Internet of Things} (IoT), the demand for deploying neural networks (NNs) on resource-constrained devices such as microcontrollers continues to grow.
For example, neural networks can be used in electric vehicles to predict battery charge~\autocite{Petersen_2022} or implement thermal management of the electric motor~\autocite{Kirchgassner_2023}.
In recent years, several mature end-to-end ML frameworks (e.g., TensorFlow~\autocite{tensorflow2015}, PyTorch~\autocite{paszke_2017}, Keras~\autocite{Chollet_2015_keras}) have emerged centered around GPUs as the workhorse. 
As a consequence, many neural network models come with high memory requirements and high computational complexity. 
This severely hampers their deployment on microcontrollers. 
Thus, compression must be used to reduce the memory footprint and speed up the computation, which typically comes at the cost of the neural network model's reduced prediction quality (e.g., accuracy, area under the curve).

A common compression technique is sparsification (a.k.a pruning~\cite{Han_2015b}), which removes parts of the neural network to reduce computational complexity and, correspondingly, inference time.
During the development of a neural network model, a programmer or data scientist is typically unaware of whether the implementation of the model will meet given execution time and memory constraints on some microcontroller target.
There is a vast design space, especially when it comes to model compression.
For example, when considering layer-wise pruning, thousands of possible combinations of pruning rates for the different layers with trainable weights exist.
Typically, this leads to time-consuming trial and error, as the model must be compressed, re-trained, and deployed multiple times on the target device until finally meeting time and memory constraints on the target hardware while still achieving a high prediction quality.

In this work, we present a toolchain that paves the way for automated model compression and code generation to mitigate this manual trial-and-error process.
It automatically returns pruned neural network models with trade-offs between memory footprint and execution time vs. prediction quality for highly resource-constrained microcontroller targets used in automotive applications, particularly for the AURIX TriCore 3xx family.\\[.75em]
\noindent
Our main contributions are:\\[-1.5em]
\begin{itemize}
   \item OpTC, a modular toolchain for deploying various types of neural networks (e.g., MLPs, CNNs, RNNs) on AURIX TriCore microcontrollers.
         Several optimizations are performed automatically, including operator fusion and model compression.
   \item Reduction of the vast design space by a \emph{sensitivity analysis}~\autocite{Han_2015b} and exploration of the different design points (neural network compression rates) to identify trade-offs between execution time, memory requirements, and prediction quality.
   \item Evaluation of the toolchain for applications from the MLPerf Tiny benchmark~\autocite{banbury_2021_mlperf} and a dataset for predicting the temperature in an electric motor~\autocite{kirchgaessner_kaggle_2021} and showcasing OpTC when exploring these applications in terms of memory footprint (RAM and ROM), execution time, and prediction quality on an AURIX TriCore~387 microcontroller.
\end{itemize}

\noindent
The remainder of this paper is organized as follows: 
\cref{sec:fundamentals} provides fundamentals on neural network pruning, \cref{sec:related_work} discusses related work, and \cref{sec:toolchain} presents the novel toolchain approach.
\cref{sec:use_case} introduces the case study, as well as the dataset and models, which are used in the experiments (\cref{sec:evaluation}) before~\cref{sec:conclusion} concludes.

\section{Fundamentals}
\label{sec:fundamentals}

Neural networks consist of multiple layers of different types and can be represented as data flow graphs.
NNs have trainable layers with weight tensors, such as fully connected and convolutional layers.
Typically, non-linear activation functions are applied after each layer, e.g., \emph{rectified linear unit} (ReLU), \emph{hyperbolic tangent} (tanh), or \emph{sigmoid} functions.
For efficient deployment of neural networks on microcontrollers, pruning~\autocite{Han_2015} has emerged as one of the most common compression techniques, which is explained below.

\subsection{Pruning}\label{sec:Pruning}
Pruning reduces the number of neurons and their connections, thereby also reducing the number of weights and the number of floating point operations (FLOPs).
\emph{Structural pruning} is a technique where entire structures, e.g., output neurons, are set to zero and can be removed from computation~\cite{Deutel_2022}.
For example, entire filters can be removed from convolutional layers, or rows and columns of the weight matrix can be eliminated in fully connected layers.
This approach can effectively decrease execution time by reducing the number of loop iterations required to process the layers.
Let a neural network be given that consists amongst other layers (e.g., activation and pooling layers) of a set $V$ of layers with trainable weights.
In the case of \emph{global pruning}, a pruning rate $p$ is defined with $0 \leq p < 1,\ p  \in \mathbb{R}$.
Further, let $M_i$ denote the number of output neurons (or filters) of each
corresponding layer $v_i \in V$.
Then, the resulting number $m_i$ of output neurons after pruning layer $v_i \in V$ is
\begin{equation}
    m_i = \lceil M_i \cdot (1-p) \rceil.
\end{equation}
In the case of \emph{layer-wise pruning}, each layer $v_i$ is assigned a pruning rate $p_i$ where $0 \leq p_i < 1,\ p_i \in \mathbb{R}$.
Typically, layer-wise pruning results in higher compression than global pruning for achieving a similar prediction quality.
Here, the resulting number $m_i$ of output neurons after pruning layer $v_i \in V$ is
\begin{equation}
    m_i = \lceil M_i \cdot (1-p_i) \rceil.
\end{equation}
The higher the pruning rate ($p$ or $p_i$), the fewer weights have to be stored, and the fewer FLOPs are required.
There exist several heuristics to determine which weights or weight structures should be removed.
The most commonly used techniques are the $\ell^1$ or $\ell^2$ norm of the weights~\cite{Heidorn22_HAEFP}.
Here, the filters (or rows and columns) with the lowest $\ell^1$ norm values are set to zero and removed.

\subsection{Design Space}\label{sec:Pruning_DSE}
When applying global pruning to a given neural network model, the size of the design space $\Omega_{\text{global}}$ of all possible pruning options is the maximal number of output neurons and filters $M_i$, which results in $\abs{\Omega_{\text{global}}} = \max_{i=0}^{\abs{V} - 1}{M_i}$. 
In the case of layer-wise pruning of a given neural network model, the size of the design space $\Omega_{\text{layer\_wise}}$ of all possible pruning combinations is thus $\abs{\Omega_{\text{layer\_wise}}} = \prod_{i=0}^{\abs{V} - 1}{M_i}$.
However, for many neural networks, this design space is excessively large, so it would take a prohibitively long time to explore and evaluate all design points.
For example, consider the \emph{Autoencoder} model for anomaly detection from the MLPerf Tiny benchmark~\cite{Banbury_2021}.
The Autoencoder contains ten fully connected layers ($\abs{V}=10$), where the number of output neurons ranges from $M_i = 8$ to $128$; then, the number of possible pruned configurations is nearly $\abs{\Omega_{\text{layer\_wise}}}=10^{20}$.
Assessing the prediction quality and inference time for all configurations is not feasible since evaluating one pruned AE model requires in our experiments approximately 40 seconds (see \cref{ssec:ae}).
The evaluation time may even increase further if retraining is performed for each pruned configuration, which might be necessary to maintain the prediction quality of a pruned neural network.

\section{Related Work}
\label{sec:related_work}

Conventional techniques for neural network deployment often ignore tight computational and memory constraints, which hinders their application to highly resource-con\-strained microcontrollers.
Therefore, a dedicated approach is required when developing and compiling neural networks for resource-constrained microcontroller targets.
It has to ensure that the computational and latency budget is within the device limits while still achieving a desired performance~\autocite{Saha_2022}.
Recently, approaches from industry and academia have emerged that differ in their support of neural network types, input formats, and support of microcontrollers as summarized in the following.

\subsection{Workflows for Deploying Neural Networks on Microcontrollers}\label{ssec:toolchains}
Software development for microcontrollers is typically based on C or C++ programming.
Thus, some ML workflows already come with an inference library \autocite{Deutel_2022,Lin_2020,David_2021,jianjia_2020} developed in C or C++.
Here, operator function calls required to compute the neural network layers are baked into C or C++ during code generation.
Typically, those lightweight inference libraries are implemented in C or C++ combined with a conversion workflow that generates library function calls for a neural network.
These workflows are often realized directly within popular machine learning (ML) frameworks, e.g., TensorFlow~\autocite{tensorflow2015} or PyTorch~\autocite{paszke_2017}, or relying on standard exchange format descriptions, such as ONNX (Open Neural Network Exchange,~\autocite{Bai_2019}) to ensure compatibility.
Typically, the inference library is intended to be sufficiently generic to be used on any 32-bit microcontroller and, therefore, portable.
Workflows such as TensorFlow Lite Micro~\autocite{David_2021} and microTVM~\autocite{Chen_2018_TVM} have drawbacks as they rely on an interpreter to execute the network graph at runtime.
This adds memory and latency overhead~\autocite{Lin_2020}.
In addition, optimizations are performed merely at the layer level of a neural network, which misses the potential of globally optimizing the overall neural network graph, e.g., by layer fusion~\autocite{Alwani_2016,Heidorn20}.
Another drawback of inference libraries is the effort required to integrate them into other standardized automotive architectures, such as AUTOSAR (Automotive Open System Architecture)~\autocite{Bunzel_2011}, a software architecture for automotive systems, especially when managing multiple dependencies and ensuring compatibility with other parts.
Within such automotive frameworks, the use of external libraries and hardware platforms (e.g., AURIX TriCore) is often restricted, necessitating a complete redesign of the inference library and workflow.

NNOM~\autocite{jianjia_2020}, TFLite Micro~\autocite{David_2021}, and DNNruntime~\autocite{Deutel_2022} are examples of generic inference libraries capable of generating ANSI C code.
TFLite Micro and DNNruntime also support the floating point format.
These libraries consist of C code that is compilable for TriCores.
However, these libraries do not consider target-specific optimizations for layers, operators, and parallel execution on multiple cores.
Implementing such target-specific optimizations (e.g., for TriCore TC3xx) would be effortful, and core-specific C code generation may be required.
One would have to strongly alter the inference libraries and add the C code for each missing operator (or layer), and usually, the code generation (i.e., for the function calls) has to be refined correspondingly.
An additional disadvantage of this approach comes when considering different optimizations. 
For example, if layer fusion shall be supported, for each combination of layers that can be fused, C code must be added to the inference library. 
As a consequence, this can add a lot of memory overhead on the target microcontroller.

Dory~\autocite{Burrello_2021} comes without an inference library and generates C code for a given model that does not require linking to an inference library.
Generating C code gives the possibility of seamless integration into other projects and provides the ability to fine-tune memory management for the target platform.
For example, one can optimize memory usage based on the specific constraints of the target microcontroller and extend code generation for the target-specific memory hierarchy.
However, Dory is dedicated to RISC-V targets and has a defined memory hierarchy.
In addition, when it comes to model compression, Dory relies on having quantized neural networks as inputs and only supports integer (int) precision.
However, we argue that on platforms such as AURIX embedded microcontrollers, which provide single-precision floating-point computation, it can still be used, especially for complex activation functions (e.g., \emph{hyperbolic tangent}), or to support mixed-precision models that are known to maintain high prediction quality.

Finally, there is a C inference library available from Ekkono\footnote{Ekkono, \enquote{From Connected To Smart}, \url{https://www.ekkono.ai/}, Date accessed: 03/19/2024} for deploying neural network models on AURIX TriCore 2xx and 3xx microcontrollers.
However, it appears that there is only a C++ inference library provided and no neural network compression is integrated into the workflow. 

\subsection{Integration of Compression Techniques into Workflows }
Typically, for compressing a given neural network, there is some trial and error phase, where, e.g., a user defines a pruning rate and measures the resulting prediction quality of the neural network on the test dataset or a subset of the test dataset.
If the model's prediction or the compression is not sufficient, this process has to be repeated several times.
Some of the workflows listed in \cref{ssec:toolchains} already integrate compression techniques.
For example, TFLite Micro is based on Tensorflow and Keras and provides quantization and weight clustering techniques.
However, for workflows such as TFLite Micro optimization options have to be set manually and the use of different techniques is only possible to a very limited extent~\autocite{Novac_2021}. 
This is a major shortcoming, as the developer will only find out if the developed neural network meets the constraints of the target hardware after deployment.
This may result in a large time overhead as the neural network has to be tuned, trained, and recompiled.

MCUNet~\cite{Lin_2020} uses neural architecture search (NAS) to find networks that meet target platform constraints.
If a model that meets the target platform constraints is found, the user can be sure that this model is deployable on the platform.
However, the models have to be trained from scratch, which is time-consuming, especially when considering, that more models have to be trained in parallel, or one huge so-called ``supernet'' has to be trained to identify subnets afterward.
Typically, there exist expert-designed pre-trained neural network models that already perform well on given datasets~\autocite{Banbury_2021}. 
In this case, such a compute- and time-expensive search strategy should be avoided.

DNNruntime scales down already defined neural networks using pruning and quantization until the memory requirements are met.
The advantage here is that existing trained neural networks can be used for deployment on the target microcontroller.
However, after compression, the model's prediction quality might be reduced.
Again, the model has to be redesigned and trained, which is time-consuming.
Especially in the case of \emph{iterative pruning}, when retraining is applied after each pruning step, this process becomes time-consuming, especially for neural nets that have multiple layers.
Here, our toolchain proposed in the following determines pruning rates automatically based on sensitivity analysis and can reduce the vast design space to only a few evaluations (which can be decided by the user).

\section{The OpTC Toolchain}
\label{sec:toolchain}

\newcommand{\tikzxmark}{%
\tikz[scale=0.23] {
    \draw[-,line width=0.7,line cap=round] (0,0) to [bend left=6] (1,1);
    \draw[-,line width=0.7,line cap=round] (0.2,0.95) to [bend right=3] (0.8,0.05);
}}
\begin{figure}[b!]
    \tikzstyle{node_type_workflow} = [rectangle, rounded corners, minimum width=2cm, text width=2cm, text depth=1.4cm, text centered,draw=black]
    \tikzstyle{process} = [rectangle, rounded corners,minimum width=2cm, minimum height=.5cm, text centered,
    text width=2cm, draw=black]
    \tikzstyle{arrow} = [thick,->,>=stealth]
    \tikzstyle{reverse_arrow} = [thick,<-,>=stealth]

    \begin{center}
    \begin{tikzpicture}[node distance=0.2cm, scale=0.1]
        \def\shiftx{2.5}
        \node (start) {};
        \node[process, right of=start, xshift=2cm,yshift=0cm,text width=1.8cm] (sensitivity) {Sensitivity analysis (Algo.~1)};
        \node[process,right of=sensitivity, minimum height=1cm, xshift=3.7cm] (pinit) {Determine initial pruning rates (Eq.~(\ref{eq:pinit}))};
        \node[process, minimum height=1cm, right of=pinit, xshift=3.4cm,text width=1.8cm] (compress) {Resulting design space $\Omega_{\text{GWP}}$}; 
        \node[process,below of=compress, minimum height=.5cm, yshift=-2cm,text width=1.4cm] (constraints) {$j \leq J$};
        \node[process, minimum height=1cm, left of=constraints, xshift=-2.5cm] (it_pruning) {Update each $p_i$ and prune~$X$ (Algo.~2)}; 
        \node[left of=it_pruning,process, xshift=-3.5cm, text width=1.7cm] (convert) {Code generation (Sect.~\ref{ssec:codegen})};

        \node[left of=convert,node_type_workflow, xshift=-2.5cm, yshift=0cm] (compile) {Compilation \& \\ deployment};
        \def\width{.5}
        \node (dec1) [rectangle, below of=compile, yshift=-0.2cm,draw=black,rounded corners,
        minimum width=0.5cm, minimum height=0.5cm] (MCU) {$\mu$C};
        \foreach \source in {60,90,120}
            \draw (MCU.\source) -- ++(0,\width);
        \foreach \source in {-30,0,30}
            \draw (MCU.\source) -- ++(\width,0);
        \foreach \source in {240,270,300}
            \draw (MCU.\source) -- ++(0,-\width);
        \foreach \source in {150,180,210}
            \draw (MCU.\source) -- ++(-\width,0);
        \node[below of=it_pruning, xshift=-1.8 cm, yshift=-2.5cm,] (result) {\scalebox{0.3}{\input{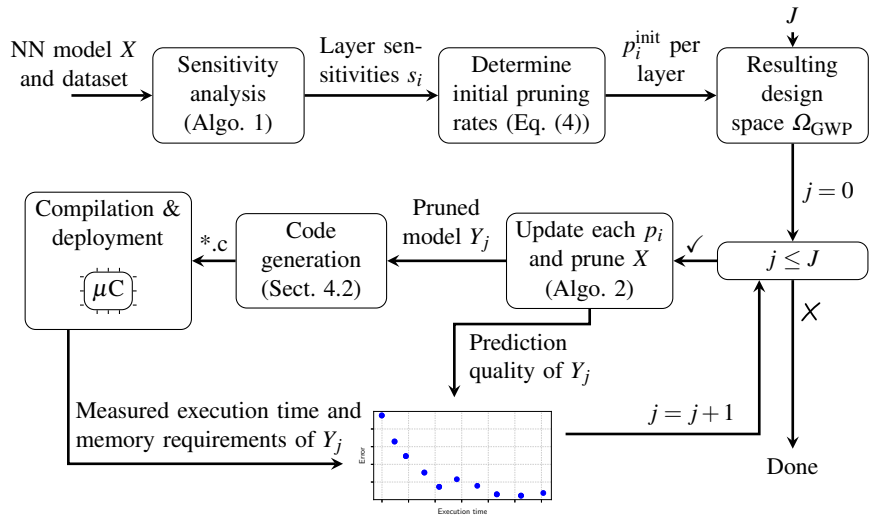}}};

        \draw [arrow,line width=1pt,draw=black] (start) -- (sensitivity) node [midway, above,text width=2.5cm, align=center, xshift=-0.5cm] {NN model $X$ and dataset};
        \draw [arrow,line width=1pt,draw=black] (sensitivity) -- (pinit) node [midway, above,text width=2.4cm, align=center] {Layer sensitivities $s_i$};
        \draw [arrow,line width=1pt,draw=black] (pinit) -- (compress) node [midway, above,text width=1.7cm, align=center] {$p^\text{init}_i$ per layer};
        \draw [reverse_arrow,line width=1pt] (compress.90) -- +(0,2) node[above] {$J$};
        \draw [arrow,line width=1pt] (compress) -- (constraints) node [midway, right] {$j=0$};
        \draw [arrow,line width=1pt] (constraints) -- (it_pruning) node [midway, above] {$\checkmark$};

        \draw [arrow,line width=1pt] (it_pruning) -- (convert) node [midway, above, text width=2cm,align=center] {Pruned model $Y_j$};
        \draw [arrow,line width=1pt] (convert) -- (compile) node [midway, above] {*.c};;
        \draw [arrow,line width=1pt] (it_pruning.270) -- +(0,-2) -| (result) node [midway, below, xshift=1.2cm,text width=2cm] {Prediction quality of $Y_j$};
        \draw [arrow,line width=1pt] (compile.240) |- (result) node [midway, above, xshift=2.2cm, text width=4.2cm] {Measured execution time and memory requirements of $Y_j$};

        \draw [arrow,line width=1pt] (result.15) -| (constraints.210) node [midway, above,xshift=-0.9cm] {$j = j+1$};
        \draw [arrow,line width=1pt] (constraints) -- +(0,-25) node[below] {Done} node [midway, right, yshift=.7cm] {\tikzxmark};
        \end{tikzpicture}
        \vspace{-.5em}
        \caption{Overview of our proposed approach to automatic compression by exploration of network configurations by iteratively increasing the degree of weighted global pruning of a neural network $X$ for deployment on the target microcontroller ($\mu$C).
        }\label{fig:workflow}
    \end{center}
    \end{figure}

According to~\cref{fig:workflow}, a sensitivity analysis is performed (\cref{alg:sensitivity}). Based on the pruning sensitivities of each layer $v_i$, initial pruning rates $p_i^\text{init}$ are determined.
Afterward, the design space is constrained by setting an upper bound $J \in \mathbb{N}$ of increasingly pruned versions of a given network model $X$.
For each pruned model $Y_j$, the prediction quality on the test dataset, the execution time, and memory requirements when deployed on the target microcontroller are determined.
The prediction quality, execution time, and memory requirements are collected, and Pareto-optimal solutions are finally output.
Our approach integrates a fully automated loop consisting of neural network (NN) compression (\cref{ssec:compression}), code generation (\cref{ssec:codegen}), flashing, and returning the measured memory utilization and execution time for the target architecture into one toolchain named OpTC.
It is thus no longer necessary for developers to perform this cycle manually and search for the optimal tradeoff between pruning rates and NN prediction quality in a trial-and-error fashion.
The toolchain supports various Python frameworks for specifying neural networks, including PyTorch and TensorFlow by relying on the ONNX standard~\autocite{Bai_2019}.
ONNX is an exchange format describing neural networks as dataflow graphs, where each node represents a mathematical operation (e.g., element-wise addition or matrix multiplication).
OpTC encompasses a template-based code generator that translates pre-trained neural networks stored in ONNX format to C code exploiting the static properties of trained neural networks (i.e., fixed layer configurations and parameters), such that no dependence on an inference library is required.
For conversion from pre-trained Pytorch or TensorFlow neural networks to ONNX, the toolchain integrates the open-source conversion tools, tf2onnx, and the PyTorch-integrated ONNX exporter.
This not only reduces the computational overhead at runtime but also allows for seamless integration into other projects or frameworks.
In the following, the steps and algorithms shown in~\cref{fig:workflow} are described in detail.
\subsection{Automatic Compression}\label{ssec:compression}
Typically, the workflow starts as follows:
A user provides the model and the dataset as well as a search space for the pruning configuration as input.
Each model is assumed to be pre-trained by an ML expert.
For achieving high flexibility, different Python libraries, such as Microsoft Neural Network Intelligence (NNI)\footnote{Microsoft, \enquote{Neural Network Intelligence (NNI)}, \url{https://github.com/microsoft/nni}}, can be integrated to compress the neural network.
This provides different options for pruning a neural network $X$, resulting in a set of $J$ pruned neural network configurations, with different trade-offs between prediction quality and execution time.

\subsubsection{Sensitivity Analysis:}\label{sec:sensitivity}
Previous works~\cite{Han_2015b,Sabih_2022_MOSP,HSMSTH24} have shown that a layer could be differently important for the prediction quality of the overall neural network.  
We perform a sensitivity analysis (\cref{alg:sensitivity}) to determine the maximal pruning rates $p_i^{\text{max}}$ for each layer $v_i \in V$ in a given neural network model $X$.

\vspace{-.5cm}
\begin{algorithm}[H]
    \caption{Sensitivity analysis}\label{alg:sensitivity}
     \textbf{Input:\ \ \ } NN model $X$ with set $V$ of trained layers, test dataset, ascending sequence of\\
     \phantom{\textbf{Input:\ \ \ }} pruning rates $P= [p^{\text{min}}, \dots, p^{\text{max}}]$, threshold $T$\\
     \textbf{Output:} Layer sensitivities $S$, i.e., $s_i \in S$ for each layer $v_i$
    \begin{algorithmic}
    \For {$v_i \in V$}
        \For{$p \gets p^{\text{min}}$ \textbf{to} $p^{\text{max}}$}

            \State {$Y \gets$ prune layer $v_i$ of model $X$ with pruning rate $p$}

            \State {$a$ $\gets$ obtain prediction quality of pruned model $Y$ by evaluating it with given test dataset}

            \If {$a < T$}
                \State {$s_i = 1-p$}  \Comment maximal pruning rate $p_i^{\text{max}}$ of layer $v_i$ found

                \State {\textbf{break}}
            \EndIf
        \EndFor
    \EndFor
    \end{algorithmic}
\end{algorithm}
\vspace{-.5cm}
For analysis of sensitivity, we define a sequence of pruning rates, e.g., $P = [0.1, 0.2,$ $ \dots, 0.9 ]$.
For each layer $v_i$, the pruning rate $p \in P$ is increased while the other layers stay unpruned and the prediction quality of the overall neural network is measured.
The sensitivity analysis usually requires just iterating over the test dataset without retraining. 
If the prediction quality $a$ of the partially pruned neural network model drops below a given threshold $T$ (e.g., the accuracy of the unpruned model), the pruning rate applied at this point defines the maximum pruning rate $p_i^{\text{max}}$ for layer $v_i$\footnote{Note that for some datasets, the prediction quality of a neural network is determined by measuring an error (e.g., the mean squared error). In this case, $a < T$ is replaced by $a > T$ in \cref{alg:sensitivity}, where $a$ is the prediction error of the pruned model, and $p_i^{\text{max}}$ is determined if the error is above the threshold error $T$ of the unpruned model.}.
The layer sensitivity $s_i$ is then defined as
\begin{equation}
    s_i = 1 - p_i^{\text{max}}.\label{eq:sensitivity}
\end{equation}
\vspace{-1cm}

\subsubsection{Global Weighted Pruning (GWP):}\label{sec:igwp}

For a given layer $v_i$, the initial pruning rate $p_i^{\text{init}}$ is determined based on the layer sensitivity $s_i$ and by introducing a positive integer number $J$ called steps, which defines how many differently pruned network configurations to be evaluated:
\begin{equation}
    p^\text{init}_i = \frac{1-s_i}{J} = \frac{p_i^{\text{max}}}{J}\qquad \forall i\ :\ v_i \in V\label{eq:pinit}
\end{equation}
Note that if $s_i=1$, $p^\text{init}_i$ turns to zero, such that the respective layer $v_i$ will stay unpruned.
We call this technique \emph{global weighted pruning}, short GWP, where each layer $v_i \in V$ gets assigned the initial pruning rate $p^\text{init}_i$, which is incremented in each iteration $j$, $0 \leq j \leq J$ that is input to \cref{alg:igwp}.
\vspace{-.5cm}
\begin{algorithm}[H]
    \caption{Global Weighted Pruning}\label{alg:igwp}
     \textbf{Input:\ \ \ } NN model $X$ with set $V$ of trained layers, initial pruning rate $p_i^{\text{init}}$ per layer $v_i$,\\
     \phantom{\textbf{Input:\ \ \ }} iteration $j$ with $0\leq j \leq J$\\
     \textbf{Output:} pruned model $Y_j$
    \begin{algorithmic}
    \For {$i \gets 0$ \textbf{to} $\abs{V}-1$}

        {$p_i = p_i^{\text{init}} \cdot j$} \Comment Determine pruning rate $p_i$ of layer $v_i$

    \EndFor\\
    $Y_j \gets$ compress model $X$ by pruning each layer $v_i$ with pruning rate $p_i$

    \end{algorithmic}
\end{algorithm}
\vspace{-.5cm}
By definition, $j=0$ denotes the initial (unpruned) neural network model $X$.
Compared to~\cite{Heidorn22_HAEFP}, where a global step size for each layer was defined, $p^\text{init}_i$ can also be interpreted as a layer-specific step size, which is kept constant, and the pruning rate $p_i$ is increased in each step $j$ by $p^\text{init}_i$.
The number of remaining filters $m_i$ after pruning for each iteration $j$ is obtained as
\begin{equation}
    m_i = \lceil M_i \cdot (1-p^\text{init}_i\cdot j) \rceil.\label{eq:mi_iterative}
\end{equation}
The design space is, therefore, of size $\abs{\Omega_{\text{GWP}}}=J$, where $J$ can be freely chosen.

\subsection{Template-based C Code Generation}~\label{ssec:codegen}

One way to compile an ONNX graph into an executable is to translate each layer directly into C code.
However, it is beneficial to optimize the graph itself, e.g., by fusing the operation nodes of the graph, as C or C++ compilers typically do not perform these optimizations~\autocite{Rotem_Glow_2018}.
Our code generator~\cite{VEHITS_2024} includes optimization techniques for merging activation functions into preceding convolutional or matrix multiplication operations.
Another example of graph optimization is the transformation of a matrix-matrix multiplication followed by a vector addition into a general matrix multiplication (GEMM), which is often required for multi-layer perceptrons, as it seems that the built-in exporters in PyTorch often model a linear layer as two separate operations (matrix-matrix multiplication with the weights, and vector addition on the resulting output by a bias).
In addition, some nodes can be removed from the computational graph of a neural network (e.g., zero-padding).
Currently, OpTC supports convolutions, linear transformations, pooling operations, and activation functions.
It also supports complex operators such as long short-term memory cells and applies approximations to activation functions.
Internally, a Python-based intermediate representation (IR) is used, which describes the neural network operators and their interactions (dataflow).
During code generation, the IR is traversed in a valid order, and for each operator (e.g., GEMM), a corresponding predefined, parameterized C template is instantiated and code emitted.
OpTC applies code optimizations (e.g., operator or loop fusion) to reduce execution time and memory requirements.
The toolchain recognizes patterns (e.g., convolution followed by ReLU), fuses them into functionally equivalent operators, and emits fused C loops, which are beneficial concerning performance.
For the intermediate tensors, tensor unionization~\cite{VEHITS_2024} is performed to wrap the tensors into unions to help the compiler reuse heap memory.

\section{Benchmarks}
\label{sec:use_case}

Neural networks have already proven their value for classification tasks (e.g., in image processing), and many works to compress them have been proposed.
However, in the case of detecting anomalies or thermal monitoring, the neural networks are typically designed to solve a regression problem, minimizing a loss function, e.g., the Mean Squared Error between the predicted value and the ground truth.

In our experiments (\cref{sec:evaluation}), we show the effects of compression for different benchmark neural networks different benchmarks and also demonstrate that our toolchain supports a broad range of different models.
For benchmarking, we use datasets and models provided by the \emph{MLPerf Tiny} benchmark suite, an open-source benchmark suite for TinyML systems~\cite{Banbury_2021}, and a publicly available \emph{Electric Motor Temperature}~(EMT) dataset from~\autocite{kirchgaessner_kaggle_2021}, consisting of temperature sequence data at different positions in an electric motor taken from a testbench.
The used datasets and neural network models are described below.

\subsection{Autoencoder (AE) for Anomaly Detection}

Analogously to MLPerf Tiny, we use the ``toy car'' sub-dataset from the ToyADMOS dataset~\cite{Koizumi_2019}.
For training, sound recordings of seven different regularly operating toy cars are provided, each consisting of 1000 audio samples mixing the machine sound with environmental noise.
The baseline architecture is an Autoencoder model consisting of an encoder (four fully connected layers) and a decoder (five fully connected layers).
Each part comprises fully connected layers (128 neurons) with ReLU activation.
The encoder and decoder are connected by a so-called bottleneck layer, which is also a fully connected layer with eight neurons.
The input and output layers of the model each have 640 neurons, respectively.
The model itself is not applied directly to the 10-second audio data.
The audio data is preprocessed into a log-mel spectrogram with 128 bands and a frame size of 32 ms. 
Then, the model is used repeatedly over a sliding window of five frames (hence the 640 input size), and the MSE of the resulting reconstruction error is averaged over the central 6.4 second part of the spectrogram, providing an overall anomaly score~\cite{Banbury_2021}.

\subsection{Convolutional Neural Network (CNN) on Keyword Spotting}\label{ssec:kws_model}
Recognition of specific words and brief phrases, known as keyword spotting, is one of the primary use cases of ultra-low-power machine learning.
Wake word detection is a specific case of keyword spotting wherein a detector continuously monitors for a specific word or phrase to enable a larger processor.
It requires continuous operation; therefore, low power consumption is key.
For this benchmark, we use the Speech Commands v2 dataset~\cite{Warden_2018}.
The dataset contains 30 words and a collection of background noises and is divided into training, validation, and test subsets such that any individual speaker only appears in one subset. 
For the model, we use the CNN model described in~\cite{Dai_2017} for processing raw audio data.
The CNN consists of five 1-dimensional convolutional layers, where the trained filters are convolved over the time dimension.
After each convolutional layer, there is a pooling layer and a ReLU activation layer. One fully connected layer at the end outputs a vector, where each element represents the probability for the respective class.
The CNN is trained on a subset of words for 12 output classes~\cite{Banbury_2021}.
To evaluate the accuracy, we randomly selected 1000 utterances from the speech commands for the test dataset.

\subsection{Temporal Convolutional Neural Network (TCN) on Electric Motor Temperature}\label{ssec:emt_model}
Temperature estimation tasks are necessary for electric drives in automotive and industrial automation applications.
As a motivating example, having fast and accurate estimations for the rotor temperature
helps to manufacture motors with fewer sensors while still enabling control strategies to utilize the motor
to its maximum capability.
For the training and evaluation, we use the publicly available \emph{Electric Motor Temperature}~(EMT) dataset from~\autocite{kirchgaessner_kaggle_2021}, consisting of temperature sequence data at different positions in a \emph{Permanent Magnet Synchronous Motor} (PMSM) motor taken from a test bench.
A total of ten input features are used, such as the speed of the electric motor, torque, current, ambient temperature, etc.
The regression targets are the temperatures of the permanent magnet, stator yoke, stator tooth, and stator winding.
In total, the dataset consists of 185 hours of recordings and integrates 69 different profiles.
For the model, the CNN of~\cite{VEHITS_2024} is chosen.
It consists of two 1-D convolutional layers with 32 1$\times$3 kernels, ReLU layers as activation functions, and one fully connected layer with four output neurons corresponding to the four target temperatures.

\section{Experiments}
\label{sec:evaluation}

The three datasets and models introduced in~\cref{sec:use_case} have been implemented in PyTorch.
Each model is trained on the respective dataset for 100 epochs, using the Adam optimizer, to minimize the Mean Squared Error (MSE) (anomaly detection and electric motor temperature), or the cross entropy (keyword spotting).
A subset of the entire dataset not used for training was used for testing.
For training and testing each explored neural network, we used an Nvidia RTX~2080~Ti GPU.
For all benchmarks, we pruned the models using the $\ell^1$ norm and setting $J=10$.
This results in 10 gradually stricter pruned model configurations.
For deployment, the PyTorch models were exported in single-precision floating-point to ONNX and converted to C code (see~\cref{ssec:codegen}).
The resulting C code was compiled using the HighTec GCC compiler\footnote{HighTec, \enquote{Tricore Development Platform v4.9.3.0-infineon-1.0}, \url{https://hightec-rt.com/en/products/development-platform}, Date accessed: 03/19/2024}, using \texttt{-O3} optimization as an argument.
Execution times were measured on the AURIX TC387\_3.3 V\_TFT evaluation board with a TriCore 387.
The TriCore 387 supports floating-point computations running at a frequency of 300 MHz.
It consists of 4 cores (CPU0, CPU1, CPU2, CPU3), which differ in the size of the scratchpad RAM, i.e., the size of scratchpad RAM of CPU0 and CPU1 is 240~kilobytes, whereas CPU2 and CPU3 are 96~kilobytes\footnote{Infineon, \enquote{AURIX TC38x User Manual}, \url{https://www.infineon.com}, Date accessed: 03/19/2024}.
The TC387 has a ROM of size 10~megabytes, where the weights of the model and the program are stored.
For a fair comparison, all models were run on a single core, with 240~kilobytes of scratchpad RAM, and the floating-point format was chosen.

\begin{figure}[b!]
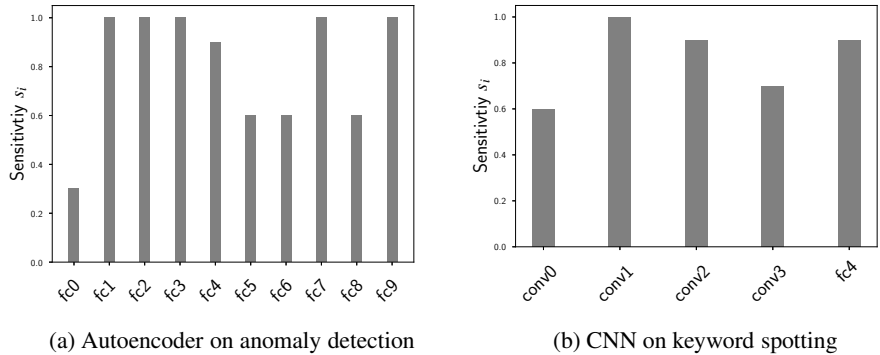

    \begin{subfigure}{0.5\textwidth}
        \scalebox{0.35}{\input{sensitivity_AE.pgf}}
        \subcaption{Autoencoder on anomaly detection}\label{sfig:AE_sensitivity}
        \end{subfigure}
        \begin{subfigure}{0.5\textwidth}
        \scalebox{0.35}{\input{sensitivity_KWS.pgf}}
        \subcaption{CNN on keyword spotting}\label{sfig:KWS_sensitivity}
        \end{subfigure}
    \caption{Pruning sensitivities $s_i$ for respective layer $v_i$, which serves to determine the initial pruning rate. The sensitivities for the ten fully connected layers of the Autoencoder (a) and for the four convolutional layers and the fully connected layer of the convolutional neural network for keyword spotting (b) are visualized.}\label{fig:sensitivity}
\end{figure}

\subsection{Anomaly Detection}\label{ssec:ae}

In the first stage of the toolchain, the sensitivity analysis is conducted for the considered model on the respective test dataset.
As a measure of the accuracy, we consider the area under the curve (AUC), which is the quality target for the anomaly detection dataset in the MLPerf tiny benchmark suite~\autocite{Banbury_2021}.
When the AUC exceeds the threshold of the unpruned model ($T=0.86$) on the test dataset (``anomaly\_id\_01''), the maximum pruning rate $p_i^{\text{max}}$ of the layer is determined (see~\cref{alg:sensitivity}).

In \cref{sfig:AE_sensitivity}, the layer sensitivities (see \cref{eq:sensitivity}) for the anomaly detection model are reported.
Remarkably, the first fully connected layer (fc0) has the lowest sensitivity, meaning that, at a maximum pruning rate of $p_0^{\text{max}}=0.7$, still the AUC of the unpruned model can be achieved.
The other layers of the encoder network (fc0 to fc3) have high sensitivities and, the fully connected layers fc1 to fc3 have to stay unpruned in the following exploration loop. 
The decoder (fc5 to fc9), has high sensitivity at the output layer, which is typical for neural networks~\autocite{Han_2015b}.
However, fc7 is an interesting case, which is highly sensitive compared to the other three layers, which is an interesting finding for this type of model.
Based on the sensitivities $s_i$, and setting $J=10$, according to \cref{eq:pinit}, the initial pruning rates $p^\text{init}_i$ for each layer are determined.
As example for fc0 $p^\text{init}_0 = \frac{1-s_0}{J} = \frac{1-0.3}{10}$. 
With $M_0=128$ output neurons, for $j=1$, the number of output neurons $m_0$ computes to $m_0 = \lceil M_0 \cdot (1-p^\text{init}_0\cdot j) \rceil =  \lceil 128 \cdot (1-\frac{1-0.3}{10}\cdot 1) \rceil =120$, meaning the $8$ output neurons with the lowest $\ell^1$ norm are pruned.
For each pruned model $Y_j$, C code is generated and the execution time as well as the memory utilization (ROM and scratchpad RAM) on the TriCore~387 are measured.

\begin{figure}[b!]
    \begin{center}
    \scalebox{0.7}{\input{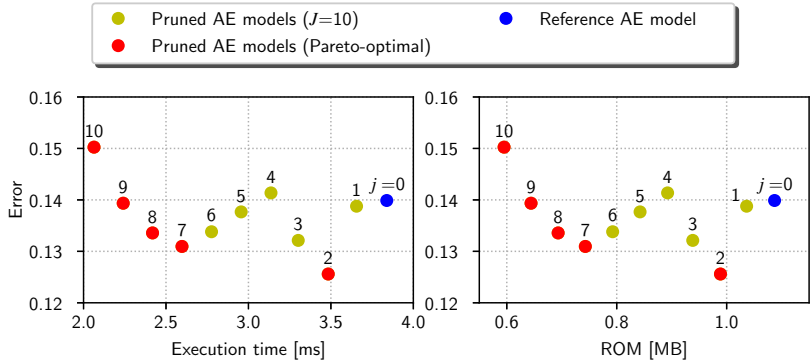}}
    \vspace{-.75em}
    \caption{$J$ explored configurations based on global weighted pruning for the Autoencoder (AE) on anomaly detection benchmark. Each dot is numbered with the corresponding iteration $j$ ($0\leq j \leq 10$).}\label{fig:AE}
    \end{center}
\end{figure}

The results are depicted in \cref{fig:AE}.
Each dot represents one explored pruned configuration.
Note that the respective y-axis denotes the error, which for the Autoencoder benchmark is the area under the curve (AUC) subtracted from 1 (Error=1-AUC).
The measurement of one pruned configuration requires about 40 seconds, with approximately 30 seconds required for determining the AUC for the test dataset and 10 seconds for C code generation, compilation, and deployment on the TC387 on the TriCore. Overall, the exploration for $J=10$ requires only 7 minutes.
The blue dot represents the unpruned baseline ($j=0$), which requires 1.1 MB ROM (11\% of the available ROM), 50 kB scratchpad RAM (21\% of available RAM of CPU0), and has an inference time of 3.8 ms.
Note that with increasing $j$ and thus increasing degree of the global weighted pruning, the number of parameters (ROM) as well as the execution time is reduced due to a reduction in the number of floating point operations required to compute the inference for the model.

The red dots represent pruned AE models that are Pareto-optimal, i.e., none of these design points is dominated by any other explored point.
Mathematically, a design point is not dominated if there is no point in the design space that is at least equally good in all objectives.
As an example, point $j=7$ dominates all points $0$ and $3-6$, as error, ROM image, and execution time are all smaller than for any of these points. 
However, $j=7$ does not dominate point $j=2$, and vice versa.

As the successive layers fc2 to fc4 stay unpruned and as these define the maximum of RAM to allocate, the RAM requirements cannot be reduced by pruning for this specific neural network model.
In \cref{fig:AE}, the non-dominated configurations $j=2$ as well as $7-9$ all dominate the initial baseline model.

Regarding the ROM requirements, OpTC also explores if a pruned network fits the memory requirements of other microcontroller targets.
For example, the TC32x microcontroller has only a ROM of 1 MB and for instance, the baseline Autoencoder model would not fit into the ROM.
Here, OpTC provides pruned configurations (e.g., all configurations $j \geq 2$ in \cref{fig:AE}) that are deployable also on TC32x microcontrollers.

\subsection{Keyword Spotting}\label{ssec:kws}

\begin{figure}[b!]
    \hspace{-3mm}
        \vspace{-.75em}
        \scalebox{0.59}{\input{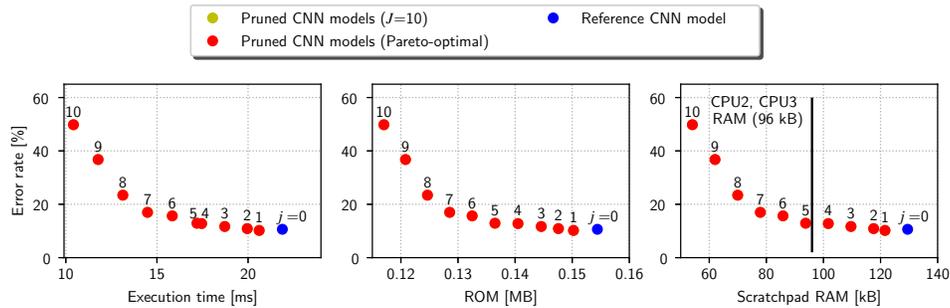}}
        \caption{$J$ explored configurations based on global weighted pruning for the convolutional neural network for the keyword spotting benchmark. The vertical line in the rightmost plot indicates the available scratchpad RAM for CPU2 and CPU3. Here, all configurations $j \geq 5$ meet the scratchpad (RAM) requirements.}\label{fig:KWS}
\end{figure}

Analogously to the Autoencoder, a sensitivity analysis of the convolutional neural network for the keyword spotting benchmark (see \cref{ssec:kws_model}) was performed.
The sensitivities in \cref{sfig:KWS_sensitivity} indicate that also the first layer (conv0) is the least sensitive.
However, compared to the Autoencoder, it still has a higher sensitivity.
Here, one can already see that the sensitivities differ significantly between different models and even an expert may have to carry out a lot of iterations to find which pruning rates for which layer yield the best trade-off between performance and prediction quality.
The number of pruning configurations to be explored is set again to $J=10$.
The error rate is computed to determine the error rate of the convolutional neural network for keyword spotting. 
It denotes the number of misclassified samples (false predicted keyword) divided by the number of samples in the test dataset.

In contrast to the Autoencoder, we see that with increasing $j$ (stricter pruning), the error rate increases as well (\cref{fig:KWS}).
Here, only configuration $j=1$ dominates the baseline, as typically pruning a network with low pruning rates shows slightly lower error rate due to better generalization\footnote{In terms of neural networks, \emph{generalization} means that a neural network tends to overfit the training dataset. With methods such as pruning, typically the overfitting on the training data can be reduced, and neural networks show better prediction quality on the test dataset.}.

Considering the model's RAM utilization, the baseline model requires 130~kB RAM, which is 54\% of the available RAM for CPU0, CPU1, respectively.
However, the baseline CNN for keyword spotting exceeds the available RAM of CPU2 and CPU3 of TC387 (96~kB, marked by the black vertical line in \cref{fig:KWS}).
Here, each of the determined pruned configurations requires only 94~kB, which fits into CPU2 and CPU3, while the prediction error increases marginally by 2.7\%.
Based on the results of OpTC, the user can decide on which CPU of the TC387 the model should be deployed and if the resulting trade-off in error still meets his constraints.

\subsection{Electric Motor Temperature}\label{ssec:emt}

\begin{figure}[b!]
\hspace{-2.5mm}
    \scalebox{0.5}{\input{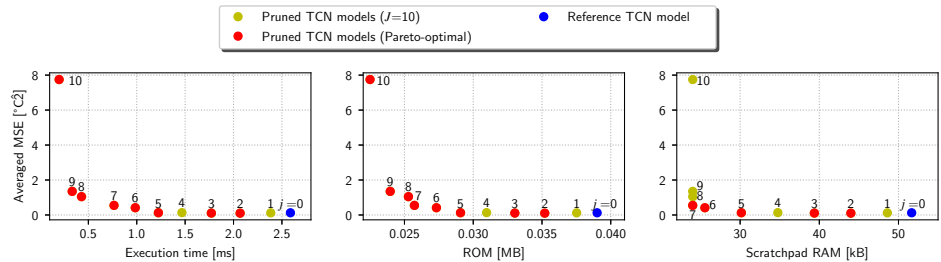}}
    \vspace{-.75em}
    \caption{$J$ explored configurations based on global weighted pruning for the temporal convolutional neural network (TCN) trained on the electric motor temperature dataset.}\label{fig:EMT}
\end{figure}

For evaluating the prediction quality of the temporal convolutional neural network for electric motor temperature prediction (see~\cref{ssec:emt_model}), we evaluated the average mean squared error (MSE) of four target temperatures.

The sensitivity analysis revealed that only the last fully connected layer is highly sensitive; Therefore, the layer stays unpruned during the exploration.
As in~\cite{VEHITS_2024}, we applied retraining of 50 epochs for each pruned model.
In \cref{fig:EMT}, the key performance indicators (i.e., execution time, MSE, and memory utilization) of the resulting pruned configurations are shown.
Also in this benchmark, one can see that with increasing $j$ and thus stricter prunings, the execution time decreases as well as ROM requirements due to the reduction of the number of parameters (and FLOPs).
The baseline model requires 2.6~ms for inference and about 52~kB of RAM.
With retraining, the mean-squared error does not increase in the first steps.
Compared to the unpruned baseline, a pruned configuration with the same MSE with an execution time of 1.2~ms was found, resulting in a speedup of $2.2\times$.
Notably, for the RAM, after steps $j=7$, no further reduction could be achieved.
At this point, the input and output features of the fully connected layer, which is left out from pruning, define the maximum of intermediate values to be stored.

\section{Conclusion}\label{sec:conclusion}
In this work, we presented OpTC, a toolchain for automated neural network model compression and C code generation for AURIX TriCore microcontrollers.
Our toolchain reduces the vast design space of pruning configurations by sensitivity analysis and is able to explore the reduced design space within minutes.
For applications from the MLPerf Tiny benchmark on TC3xx targets that are commonly used in automotive systems, we showcased the efficiency of OpTC, providing speedups or reducing the memory footprint to make models deployable also on targets with a small capacity of scratchpad (RAM) memory.
In the case of the anomaly detection and electric motor temperature benchmarks, OpTC provides pruned configurations showing speedups of $1.7 \times$ and $2.2\times$, respectively, without increasing the error over the baseline model.

Furthermore, we showed that the explorations using OpTC also enable increasing the range of microcontroller targets to which a given neural network can be deployed.
For example, the TC32x microcontroller has only a ROM size of 1~MB.
Here, the baseline Autoencoder model from the MLPerf Tiny Benchmark would not fit into the ROM.
OpTC provides pruned configurations that are deployable on TC32x microcontrollers.
In the case of the keyword spotting benchmark, we demonstrated how OpTC determines configurations that can be run on cores with reduced RAM size.

\vspace{1em}
{
\noindent\textbf{Acknowledgements:\ } This work was supported by the Schaeffler Hub for Advanced Research at Friedrich-Alexander-Universität Erlangen-Nürnberg (SHARE at FAU).}

\printbibliography

@string { InProc = "In "}

@article{Bunzel_2011,
  author       = {Stefan Bunzel},
  title        = {{AUTOSAR} -- {The} Standardized Software Architecture},
  journal      = {Informatik Spektrum},
  volume       = {34},
  number       = {1},
  pages        = {79--83},
  year         = {2011},
  doi          = {10.1007/S00287-010-0506-7}
}

@inproceedings{Dai_2017,
  author       = {Wei Dai and
                  Chia Dai and
                  Shuhui Qu and
                  Juncheng Li and
                  Samarjit Das},
  title        = {Very deep convolutional neural networks for raw waveforms},
  booktitle    = {Proceedings of the IEEE International Conference on Acoustics, Speech and Signal Processing (ICASSP)},
  venue ={New Orleans, LA, USA},
  date={2017-03-05/2017-03-09},
  pages        = {421--425},
  publisher    = {IEEE},
  doi          = {10.1109/ICASSP.2017.7952190}
}

@article{Warden_2018,
  author       = {Pete Warden},
  title        = {Speech Commands: {A}{} Dataset for Limited-Vocabulary Speech Recognition},
  year         = {2018},
  eprint       = {1804.03209},
  journal      = {The Computing Research Repository (CoRR)},
  eprinttype   = {arxiv},
  eprintclass  ={cs.CL}
}

@misc{Chollet_2015_keras,
  title={Keras},
  author={Chollet, Fran\c{c}ois and others},
  year={2015},
  howpublished={\url{https://keras.io}},
}

@inproceedings{Koizumi_2019,
  author       = {Yuma Koizumi and
                  Shoichiro Saito and
                  Hisashi Uematsu and
                  Noboru Harada and
                  Keisuke Imoto},
  title        = {{ToyADMOS}: {A}{} Dataset of Miniature-Machine Operating Sounds for Anomalous Sound Detection},
  booktitle    = {Proceedings of the IEEE Workshop on Applications of Signal Processing to Audio and Acoustics (WASPAA)},
  venue ={New Paltz, NY, USA},
  date={2019-10-20/2019-10-23},
  pages        = {313--317},
  publisher    = {IEEE},
  doi          = {10.1109/WASPAA.2019.8937164}
}

@inproceedings{Banbury_2021,
  author       = {Colby R. Banbury and
                  Vijay Janapa Reddi and
                  Peter Torelli and
                  Nat Jeffries and
                  Csaba Kir{\'{a}}ly and
                  Jeremy Holleman and
                  Pietro Montino and
                  David Kanter and
                  Pete Warden and
                  Danilo Pau and
                  Urmish Thakker and
                  Antonio Torrini and
                  Jay Cordaro and
                  Giuseppe Di Guglielmo and
                  Javier M. Duarte and
                  Honson Tran and
                  Nhan Tran and
                  Wenxu Niu and
                  Xuesong Xu},
  editor       = {Joaquin Vanschoren and
                  Sai{-}Kit Yeung},
  title        = {MLPerf Tiny Benchmark},
  booktitle    = {Proceedings of the Neural Information Processing Systems Track on
                  Datasets and Benchmarks 1, NeurIPS Datasets and Benchmarks 2021, December
                  2021, virtual},
  year         = {2021},
  url          = {https://datasets-benchmarks-proceedings.neurips.cc/paper/2021/hash/da4fb5c6e93e74d3df8527599fa62642-Abstract-round1.html},
  bibsource    = {dblp computer science bibliography, https://dblp.org}
}

@inproceedings{Han_2015b,
  author       = {Song Han and
                  Jeff Pool and
                  John Tran and
                  William J. Dally},
  title        = {Learning both Weights and Connections for Efficient Neural Network},
  booktitle    = InProc # {Proceedings of the Annual Conference on Neural Information Processing Systems (NIPS)},
  pages        = {1135--1143},
  year         = {2015}
}

@article{banbury_2021_mlperf,
  title={MLPerf Tiny Benchmark},
  author={Banbury, Colby and Reddi, Vijay Janapa and Torelli, Peter and Holleman, Jeremy and Jeffries, Nat and Kiraly, Csaba and Montino, Pietro and Kanter, David and Ahmed, Sebastian and Pau, Danilo and others},
  journal={Proceedings of the Neural Information Processing Systems Track on Datasets and Benchmarks},
  year={2021}
}

@inproceedings{Lin_2020,
  author       = {Ji Lin and
                  Wei{-}Ming Chen and
                  Yujun Lin and
                  John Cohn and
                  Chuang Gan and
                  Song Han},
  title        = {{MCUNet}: {Tiny} Deep Learning on {IoT} Devices},
  booktitle    = InProc # {Proceedings of the Annual Conference on Neural Information Processing Systems (NeurIPS)},
  year         = {2020}
}

@inproceedings{Han_2015,
  author       = {Song Han and
                  Huizi Mao and
                  William J. Dally},
  title        = {Deep Compression: {Compressing} Deep Neural Network with Pruning, Trained Quantization and {Huffman} Coding},
  booktitle    = InProc # {Proceedings of 4th International Conference on Learning Representations (ICLR)},
  year         = {2016},
  doi={10.48550/arXiv.1510.00149}
}

@article{Kirchgassner_2023,
  author = {Wilhelm Kirchg{\"{a}}ssner and Oliver Wallscheid and Joachim B{\"{o}}cker},
  title        = {Thermal neural networks: {Lumped}-Parame\-ter Thermal Modeling with State-Space Machine Learning},
  journal      = {Engineering Applications of Artificial Intelligence},
  volume       = {117},
  pages        = {105537},
  year         = {2023},
  doi          = {10.1016/J.ENGAPPAI.2022.105537}
}

@misc{kirchgaessner_kaggle_2021,
title={Electric Motor Temperature},
url={https://www.kaggle.com/dsv/2161054},
DOI={10.34740/KAGGLE/DSV/2161054},
publisher={Kaggle},
author = {Wilhelm Kirchg{\"{a}}ssner and Oliver Wallscheid and Joachim B{\"{o}}cker},
year={2021}
}

@article{Rotem_Glow_2018,
  author       = {Nadav Rotem and
                  Jordan Fix and
                  Saleem Abdulrasool and
                  Summer Deng and
                  Roman Dzhabarov and
                  James Hegeman and
                  Roman Levenstein and
                  Bert Maher and
                  Nadathur Satish and
                  Jakob Olesen and
                  Jongsoo Park and
                  Artem Rakhov and
                  Misha Smelyanskiy},
  title        = {Glow: {Graph} Lowering Compiler Techniques for Neural Networks},
  journal      = {The Computing Research Repository (CoRR)},
  eprint={1805.00907},
  eprinttype   = {arxiv},
  eprintclass={cs.PL},
  year         = {2018}
}

@inproceedings{Petersen_2022,
  author       = {Patrick Petersen and
                  Thomas Rudolf and
                  Eric Sax},
  title        = {A Data-driven Energy Estimation based on the Mixture of Experts Method
                  for Battery Electric Vehicles},
  booktitle    = InProc # {Proceedings of the 8th International Conference on Vehicle Technology and Intelligent Transport Systems (VEHITS)},
  pages        = {384--390},
  publisher    = {{SCITEPRESS}},
  year         = {2022},
  doi          = {10.5220/0011081000003191}
}

@inproceedings{VEHITS_2024,
  author       = {{Heidorn}, {Christian} and {Hannig}, {Frank} and {Riedelbauch}, {Dominik} and {Strohmeyer}, {Christoph} and {Teich}, {Jürgen}},
  booktitle    = {Proceedings of the 10th International Conference on Vehicle Technology and Intelligent Transport Systems (VEHITS)},
  venue        = {Angers, France},
  publisher    = {SciTePress},
  title        = {Efficient Deployment of Neural Networks for Thermal Monitoring on AURIX TC3xx Microcontrollers},
  date         = {2024-05-02/2024-05-04}
}

@article{Novac_2021,
  author       = {Pierre{-}Emmanuel Novac and
                  Ghouthi Boukli Hacene and
                  Alain Pegatoquet and
                  Benoit Miramond and
                  Vincent Gripon},
  title        = {Quantization and Deployment of Deep Neural Networks on Microcontrollers},
  journal      = {Sensors},
  volume       = {21},
  number       = {9},
  pages        = {2984},
  year         = {2021},
  doi          = {10.3390/s21092984}
}

@inproceedings{Alwani_2016,
  author       = {Manoj Alwani and
                  Han Chen and
                  Michael Ferdman and
                  Peter A. Milder},
  title        = {Fused-Layer {CNN} Accelerators},
  booktitle    = InProc # {Proceedings of the 49th Annual IEEE/ACM International Symposium on Microarchitecture (MICRO)},
  venue = {Taipei, Taiwan},
  pages        = {22:1--22:12},
  publisher    = {IEEE Computer Society},
  year         = {2016},
  doi          = {10.1109/MICRO.2016.7783725},
}

@inproceedings{Chen_2018_TVM,
  author       = {Tianqi Chen and
                  Thierry Moreau and
                  Ziheng Jiang and
                  Lianmin Zheng and
                  Eddie Q. Yan and
                  Haichen Shen and
                  Meghan Cowan and
                  Leyuan Wang and
                  Yuwei Hu and
                  Luis Ceze and
                  Carlos Guestrin and
                  Arvind Krishnamurthy},
  title        = {{TVM:} {An} Automated End-to-End Optimizing Compiler for Deep Learning},
  booktitle    = InProc # {Proceedings of the 13th USENIX Symposium on Operating Systems Design and Implementation (OSDI)},
  pages        = {578--594},
  publisher    = {{USENIX} Association},
  year         = {2018}
}

@article{Deutel_2022,
  author       = {Mark Deutel and
                  Philipp Woller and
                  Christopher Mutschler and
                  J{\"{u}}rgen Teich},
  title        = {Energy-efficient Deployment of Deep Learning Applications on {Cortex-M} based Microcontrollers using Deep Compression},
  journal      = {The Computing Research Repository (CoRR)},
  year         = {2022},
  eprint       = {2205.10369},
  eprinttype   = {arxiv},
  eprintclass={cs.LG}
}

@article{Burrello_2021,
  author       = {Alessio Burrello and
                  Angelo Garofalo and
                  Nazareno Bruschi and
                  Giuseppe Tagliavini and
                  Davide Rossi and
                  Francesco Conti},
  title        = {{DORY:} {Automatic} End-to-End Deployment of Real-World {DNNs} on Low-Cost {IoT} {MCUs}},
  journal      = {IEEE Transactions on Computers},
  volume       = {70},
  number       = {8},
  pages        = {1253--1268},
  year         = {2021},
  doi          = {10.1109/TC.2021.3066883}
}

@misc{Bai_2019,
    author = {Bai, Junjie and Lu, Fang and Zhang, Ke and others},
    title = {{ONNX}: Open Neural Network Exchange},
    year = {2019},
    publisher = {GitHub},
    journal = {GitHub repository},
    url = {https://github.com/onnx/onnx}
}

@inproceedings{paszke_2017,
  title={Automatic differentiation in {PyTorch}},
  author={Paszke, Adam and Gross, Sam and Chintala, Soumith and Chanan, Gregory and Yang, Edward and DeVito, Zachary and Lin, Zeming and Desmaison, Alban and Antiga, Luca and Lerer, Adam},
  booktitle=InProc # {Proceedings of NIPS Autodiff Workshop},
  publisher    = {OpenReview.net},
  year         = {2017},
  url          = {https://openreview.net/forum?id=BJJsrmfCZ}
}

@software{jianjia_2020,
  author       = {Jianjia Ma},
  title        = {{A higher-level Neural Network library on Microcontrollers (NNoM)}},
  month        = oct,
  year         = 2020,
  publisher    = {Zenodo},
  version      = {v0.4.2},
  doi          = {10.5281/zenodo.4158710}
}

@misc{tensorflow2015,
title={{TensorFlow}: Large-Scale Machine Learning on Heterogeneous Systems},
url={https://www.tensorflow.org},
author={
    Mart\'{i}n Abadi and
    Ashish Agarwal and
    Paul Barham and others},
  year={2015},
}

@article{Saha_2022,
  author       = {Swapnil Sayan Saha and
                  Sandeep Singh Sandha and
                  Mani B. Srivastava},
  title        = {Machine Learning for Microcontroller-Class Hardware: {A} Review},
  journal      = {The Computing Research Repository (CoRR)},
  eprint={2205.14550},
  eprinttype   = {arxiv},
  eprintclass={cs.LG},
  year         = {2022}
}

@article{David_2021,
  author       = {Robert David and
                  Jared Duke and
                  Advait Jain and
                  Vijay Janapa Reddi and
                  Nat Jeffries and
                  Jian Li and
                  Nick Kreeger and
                  Ian Nappier and
                  Meghna Natraj and
                  Shlomi Regev and
                  Rocky Rhodes and
                  Tiezhen Wang and
                  Pete Warden},
  title        = {{TensorFlow Lite Micro}: {Embedded} Machine Learning on {TinyML} Systems},
  journal      = {The Computing Research Repository (CoRR)},
  year         = {2020},
  eprint       = {2010.08678},
  eprinttype   = {arxiv},
  eprintclass={cs.AI}
}

@inproceedings{Heidorn20,
  author    = {Christian Heidorn and
               Frank Hannig and
               J{\"{u}}rgen Teich},
  title     = {Design Space Exploration for Layer-parallel Execution of Convolutional Neural Networks on CGRAs},
  booktitle = {Proceedings of the 23rd International Workshop on Software and Compilers for Embedded Systems (SCOPES)},
  pages     = {26--31},
  publisher = {{ACM}},
  year      = {2020},
  doi       = {10.1145/3378678.3391878},
}

@inproceedings{Heidorn22_HAEFP,
  author    = {Christian Heidorn and
               Nicolai Meyerh{\"{o}}fer and
               Christian Schinabeck and
               Frank Hannig and
               J{\"{u}}rgen Teich},
  title     = {Hardware-Aware Evolutionary Filter Pruning},
  booktitle = {Embedded Computer Systems: Architectures, Modeling, and Simulation
               - 22nd International Conference, {SAMOS} 2022, Samos, Greece, July
               3-7, 2022, Proceedings},
  series    = {Lecture Notes in Computer Science},
  volume    = {13511},
  pages     = {283--299},
  publisher = {Springer},
  year      = {2022},
  doi       = {10.1007/978-3-031-15074-6\_18}
}

@article{HSMSTH24,
  author       = {{Heidorn}, {Christian} and {Sabih}, {Muhammad} and {Meyerhöfer}, {Nicolai} and {Schinabeck}, {Christian} and {Teich}, {Jürgen} and {Hannig}, {Frank}},
  journal      = {International Journal of Parallel Programming},
  title        = {Hardware-Aware Evolutionary Explainable Filter Pruning for Convolutional Neural Networks},
  year         = {2024},
  doi          = {10.1007/s10766-024-00760-5}
}

@inproceedings{Sabih_2022_MOSP,
  author       = {Muhammad Sabih and
                  Ashutosh Mishra and
                  Frank Hannig and
                  J{\"{u}}rgen Teich},
  title        = {{MOSP:} Multi-Objective Sensitivity Pruning of Deep Neural Networks},
  booktitle    = {13th {IEEE} International Green and Sustainable Computing Conference,
                  {IGSC} 2022, Pittsburgh, PA, USA, October 24-25, 2022},
  pages        = {1--8},
  publisher    = {{IEEE}},
  year         = {2022},
  url          = {https://doi.org/10.1109/IGSC55832.2022.9969374},
  doi          = {10.1109/IGSC55832.2022.9969374},
  bibsource    = {dblp computer science bibliography, https://dblp.org}
}

\end{document}